# Obstacle evasion using fuzzy logic in a sliding blades problem environment

Shadrack Kimutai
Information Communication Technology Department
Rift Valley Technical Training Institute
244 Eldoret Kenya
Tel: +254724226334
Shadrackkimutai@gmail.com
Submitted on 1/5/2016

**Abstract**

*This paper discusses obstacle avoidance using fuzzy logic and shortest path algorithm. This paper also introduces the sliding blades problem and illustrates how a drone can navigate itself through the swinging blade obstacles while tracing a semi-optimal path and also maintaining constant velocity.*

## Introduction

The Current technological know-how in the field of computing, robotics and navigation is sufficient for development of self-navigated drones (robotic vehicles). In the recent years drones have become popular due to its ability to operate in environments that are hazardous to a human operator. Sadly though, nearly all drones be they space probes, submersibles or aircraft drones still rely on a human controller in areas such as navigating in environments consisting of chaotically moving Obstacles such as in the asteroid belt or in deep sea diving.

Currently there are a handful of algorithms that offer good results example being the Virtual force field algorithm which has been in use for quite some time. This algorithm is based on the illusion that the robot repels the obstacle and tries any angle within $180^0$ to evade colliding with the object. While this method is extremely simple and practical in situations where the obstacle is static, it fails drastically with mobile obstacles as a result many other algorithms such as Kyongs Modified virtual force field (MVFF) (Kwon & Cho, 2006)which is a more robust algorithm capable of mapping an environment with mobile obstacles. The shortcoming of this algorithm is that it doesn't factor in the shortest path approach to reaching the destination even when the ideal path is obstructed. Most of the methods identified (let alone a few like the General approach theory(Jaafar & McKenzie, 2007)) do not factor in the consideration of the optimal path redefinition hence resulting to the trajectory being followed being selected without considerations of its cost.(Yadav & Biswas, 2010)

We will use a combination of fuzzy logic and shortest path first to establish the optimal path around obstacles between the source and the destination of the drone. To achieve this we are going to model the above scenario into the swinging blade problem which is a common test for navigation skills in gaming environments.

## Approach to the Problem

As identified above, we will use the swinging blade problem to establish our algorithm. Simply put swinging blade problem seeks to test the agent's skill in navigating the swinging obstacles without colliding with either of them. It's also important to state that the drone must have a 180⁰ "view" of the trajectory this may be possible through sensors such as sonar, edge mapping or any other technique that may deliver the same

A typical environment in the swinging blade problem looks as shown below

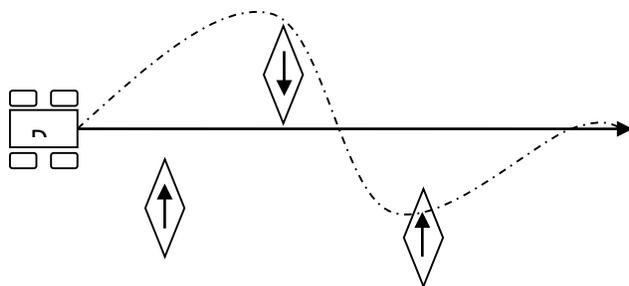

Whereby the dashed line shows the actual path followed and the bold straight line being the ideal path which also happens to be the shortest ideal path.

This can be presented as follows

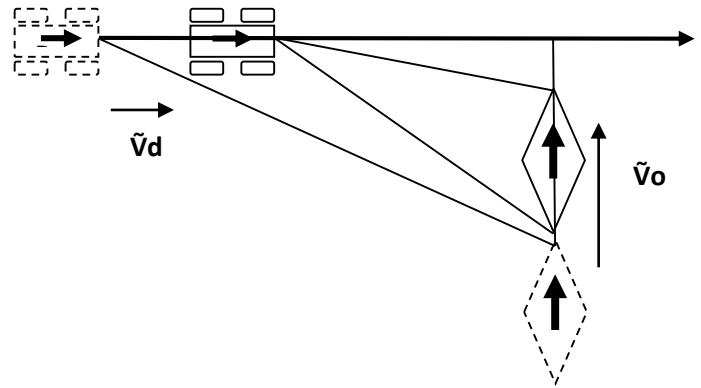

As per the diagram above the vectors $\tilde{V}d$ represents the drone path while $\tilde{V}o$ represents the blade path

Assume we draw the resultant vector diagram now that we have a clear picture of the problem is. We end up with this

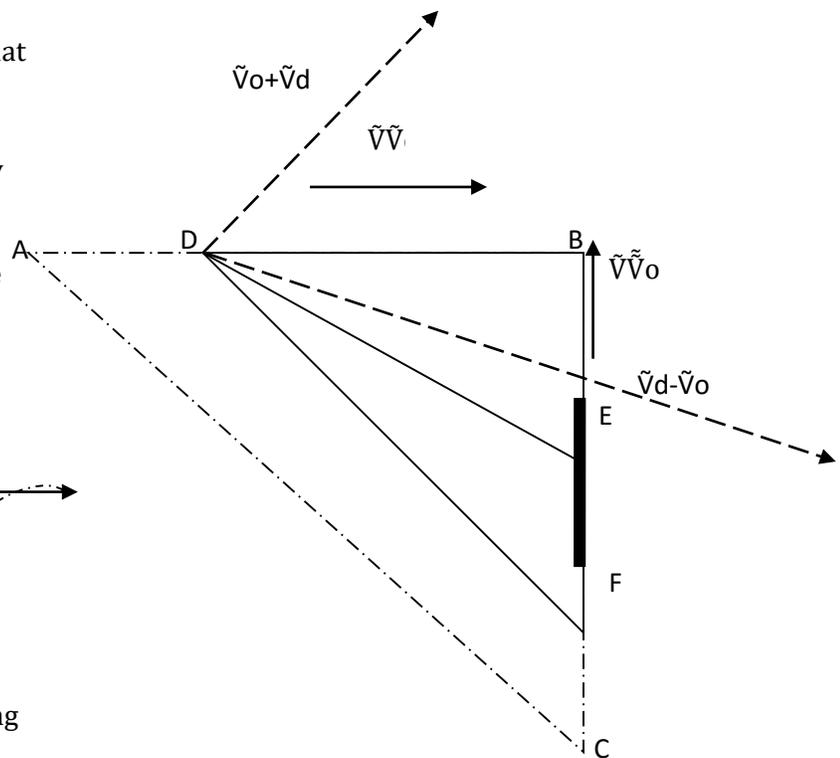

From the abstracted diagram we can deduce the size of the object based on which we draw several scenarios we will later see, furthermore, at point D the drone must immediately calculate the following

1. The velocity vector Vo of the Obstruction and the rate it is approaching drone's path.
2. Alternate paths to escape collision
3. Evaluate the cost of alternate paths
   Alternate paths include the following
   We also have to assume that velocity vectors $\tilde{V}o$ and $\tilde{V}d$ are equal
   a) if Sin BAC=Sin BDE
      favor path identified by vector $\tilde{V}o+\tilde{V}d =\tilde{v}$
   b) if Sin BAC<Sin BDE
      Maintain path along Vector $\tilde{V}d$
   c) if Sin BÃC > Sin BD̃E
      favor vector $\tilde{V}d-\tilde{V}o=-\tilde{v}$
      if and only if the size of the blade which spans the angle <EDF should be smaller than <EDB else a new vector $\tilde{\beta}$ which is directly proportional to the size of the blade has to be added to $-\tilde{v}$ above so as to end up with $-\tilde{v}+\tilde{\beta}$

We then map the above paths into our fuzzy relation as follows

$$\begin{cases} \tilde{v} \\ 0 \\ -\tilde{v} \end{cases}$$

Now we establish their path having low cost so as to associate it with the highest probability.

We let $\tilde{v}$ be 1 when if Sin BAC-Sin BDE<0.5

And $-\tilde{v}$ be zero at this juncture

We let $\tilde{v}$ and $-\tilde{v}$ be each at least 0.3 when

Sin BAC < Sin BDE

We let $\tilde{v}$ be 0 and $-\tilde{v}$ be 1 when

if Sin BÃC > Sin BD̃E

with the above rules in place we can now draw the fuzzy graph of the navigator

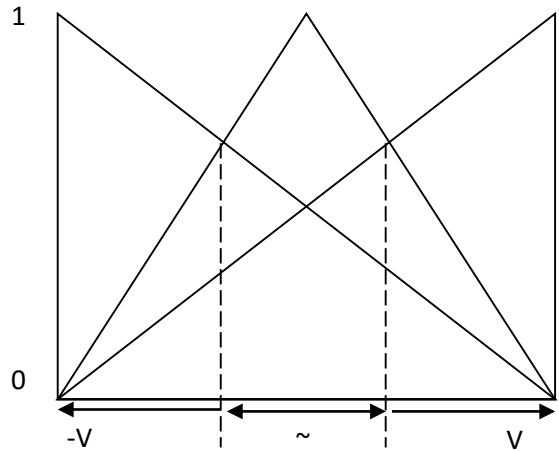

However if the size of the obstacle is large, i.e. if E⟷F is wide then $\tilde{v}$ has to be taken irrespective of if Sin BÃC > Sin BD̃E however besides this, the velocity Vector has to be increased to counter the angle difference hence the resultant vector should be

$-\tilde{v}+\alpha$

where $\alpha$ is the Boost which in the real world will involve conversion of fuel to energy

So far we have finished modeling for only one Obstacle. Now after each obstacle has been evaded, the drone may find itself away from the optimal path while other algorithms such as Virtual force field does return the drone to its initial path, we felt that this was not the best option especially if the displacement from the ideal path is large instead, a new path is drawn from the present location to the goal this trajectory will follow the hypotenuse with the separation between the ideal path and the new path spanning an angle at the goal.

The algorithm of the navigation procedure is as shown below

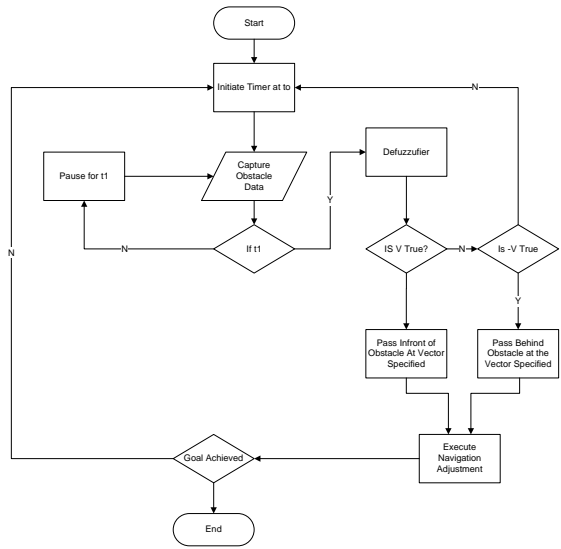

This method offers the simplest non-virtual force field approach which based on theoretical facts offers better automatic navigation compared to VFF, while we believe there is still a lot to be done in the area we have presented we feel that we have provided enough foundation on the method.

## Related work

Other authors have approached this problem differently for instance Kwon and Cho(Kwon & Cho, 2006) who in their paper modified VFF to a versatile MVFF which according to the paper was developed for submersibles and is able to track obstacles and automatically evade them. Other authors in the field are Oussama (Khatib, 1999)who in his works presents methods in which robotic mechanical parts (such as robotic arms) can move about without encountering obstruction.

## Conclusion

With the cost of fuel and any other form of energy used to power drones being limited, it's now important to ensure that the drones do not lose their momentum because of obstacles in their path at the same time it's important to ensure that the path followed is as short as possible. In this paper we have offered a fuzzy logic optimal path algorithm that allows a self-propelled and navigated drone. we have covered much of the theoretical work therefore areas that interested researchers ought to look in this topic in future include the practical application of the algorithm and evaluation of the algorithm in a swinging blade environment where the obstacles are moving in different directions other Rather than cutting across as presented in this paper another interesting area includes evaluation of this algorithm in an environment with other similar drones.

## Works Cited

Borenstein, J., & Koren, Y. (1996). Real-time Obstacle Avoidance for Fast Mobile Robots. *Systems, Man, and Cybernetics* (pp. 2-3). IEEE.

Jaafar, J., & McKenzie, E. (2007). A Fuzzy Action Selection Method for Virtual Agent Navigation in Unknown Virtual Environments. 6-7.

Khatib, O. (1999). Real time Obstacle avoidance for manipulators and Mobile Robots. *IEEE* , pp. 94-97.

Kwon, K.-Y., & Cho, J. (2006). Collision Avoidance of Moving Obstacles for Underwater Robots . *SYSTEMICS, CYBERNETICS AND INFORMATICS* (pp. 86-89). Changwon,Korea: IEEE.

Yadav, K., & Biswas, R. (2010). An Approach to Find kth Shortest Path Using fuzzy Logic. *INTERNATIONAL JOURNAL OF COMPUTATIONAL COGNITION* , 1.